\documentclass[11pt,a4paper]{article}
\usepackage{times,latexsym}
\usepackage{url}
\usepackage[T1]{fontenc}
\usepackage{xcolor}
\usepackage{colortbl}
\usepackage{graphicx}
\usepackage{caption}
\usepackage{subcaption}
\usepackage[most]{tcolorbox}
\usepackage{cuted}
\usepackage{booktabs}
\usepackage{amsmath}
\usepackage{amssymb}
\usepackage{framed}
\usepackage{multirow}
\usepackage{array}

\usepackage[table]{xcolor}
\definecolor{modelbg}{RGB}{225,225,225}
\definecolor{benchbg}{RGB}{240,240,240}

\usepackage{xurl}            
\usepackage{hyperref}
\def\paperDraft{}
\ifdefined\paperDraft
  \def\ni#1{{\color{blue}[Nikolai: \textit{#1}]}}

  \def\sharidedit#1{{\color{red}#1}}

\else
  \def\ni#1{}
  
\fi

\newcommand{\modelname}[1]{\textsc{#1}}

\definecolor{rLow}{HTML}{EFF3FF}
\definecolor{rMid}{HTML}{BDD7E7}
\definecolor{rHigh}{HTML}{6BAED6}
\definecolor{rTop}{HTML}{2171B5}

\usepackage{wrapfig}
\usepackage{ragged2e}

\definecolor{peach}{HTML}{FFDFD3}
\definecolor{rowgray}{gray}{0.95}
\definecolor{taskgray}{gray}{0.88}
    \usepackage[acceptedWithA]{tacl2021v1}
\usepackage{tacl2021v1}

\usepackage{xspace,mfirstuc,tabulary}

\newcommand{\sepmark}{\textcolor{blue}{\textbf{\textsc{[SEP]}}}}

\newif\iftaclinstructions
\taclinstructionsfalse 
\iftaclinstructions
\renewcommand{\confidential}{}
\renewcommand{\anonsubtext}{(No author info supplied here, for consistency with
TACL-submission anonymization requirements)}
\newcommand{\instr}
\fi

\usepackage{enumitem}
\usepackage{adjustbox}

\iftaclpubformat 

\else

\fi

\usepackage{tabularx,siunitx}
\title{Humans vs Vision-Language Models:\\A Unified Measure of Narrative Coherence} 

\author{
  Nikolai Ilinykh\textsuperscript{\textdagger},\,
  Hyewon Jang\textsuperscript{\textdagger},\,
  Shalom Lappin\textsuperscript{\textsection \textdagger},\,
  Asad Sayeed\textsuperscript{\textdagger},
  \and Sharid Lo\'{a}iciga\textsuperscript{\textdagger} \\
  \textsuperscript{\textdagger}Dept. of Philosophy, Linguistics, and Theory of Science, University of Gothenburg \\
  \textsuperscript{\textsection}School of Electronic Engineering and Computer Science, Queen Mary University of London \\
  \textsuperscript{\textsection}Dept. of Informatics, King's College London \\
  \texttt{nikolai.ilinykh@gu.se, hyewon.jang@gu.se, s.lappin@qmul.ac.uk,} \\
  \texttt{asad.sayeed@gu.se, sharid.loaiciga@gu.se}
}

\date{}

\begin{document}
\maketitle
\begin{abstract}
    We study narrative coherence in visually grounded stories by comparing human-written narratives with those generated by vision-language models (VLMs) on the Visual Writing Prompts corpus. Using a set of metrics that capture different aspects of narrative coherence, including coreference, discourse relation types, topic continuity, character persistence, and multimodal character grounding, we compute \textit{a narrative coherence score}. We find that VLMs show broadly similar coherence profiles that differ systematically from those of humans. In addition, differences for individual measures are often subtle, but they become clearer when considered jointly. Overall, our results indicate that, despite human-like surface fluency, model narratives exhibit systematic differences from those of humans in how they organise discourse across a visually grounded story.
    Our code is available at \url{https://github.com/GU-CLASP/coherence-driven-humans}.
\end{abstract}

\section{Introduction}

With the growing ability of large language models (LLMs) and vision-language models (VLMs) to generate fluent, human-like text, there has been growing interest in rigorously characterising the properties of automatically generated texts \citep{munoz-ortiz_contrasting_2024,zamaraeva-etal-2025-comparing}, especially in multimodal tasks such as image captioning \citep{kasai-etal-2022-transparent}.

In this context, coherence is a central property to consider, as it determines whether a text can be interpreted as a connected discourse, rather than a sequence of individually fluent but unrelated sentences. Coherence is a property of discourse that makes it intelligible, meaningful, and appropriate to its context \citep{HallidayHasan1976}. 

%

Model-generated text often differs systematically from human-written text. Recent work shows, for example, that it often follows templatic patterns \citep{namuduri-qudsim_2025} and diverges from human text in coherence-related measures such as coreference \citep{ilinykh-etal-2025-coreference}. However, these differences may be difficult for human readers to detect, because model outputs are highly fluent. In this paper, we show that this also holds for the coherence properties of visually grounded narratives.

We investigate coherence in 
stories about sequences of images, using the Visual Writing Prompts (VWP) corpus \citep{hong-etal-2023-visual-writing}. We compare human-written VWP narratives with stories generated by vision-language models (VLMs) from the same image sequences. To characterize coherence in this genre, we use five measures: coreference, implicit discourse relation typology, topic progression, character persistence, and multimodal character grounding. These measures capture complementary aspects of discourse organisation and are combined into a single scoring function, the \textbf{Narrative Coherence Score} (NCS).


Our results show that texts produced by different VLMs exhibit broadly similar coherence profiles, which differ systematically from those of human-written narratives. Although these differences are subtle at the level of individual measures, they become more pronounced when the metrics are considered jointly through the proposed narrative coherence score. Taken together, the findings suggest differences in how humans and models organize and integrate discourse information over the course of a narrative, despite comparable surface-level fluency.

\section{Related Work}

\textbf{Computational work} on discourse coherence has proposed operationalizations that capture different aspects of discourse organization. Prominent approaches include entity-based models, which track discourse participants across sentences \citep{barzilay-lapata-2008-modeling,elsner-charniak-2011-extending}, and discourse-relation frameworks, which model the semantic relations connecting discourse segments \citep{lin-etal-2011-automatically,rohde-etal-2018-discourse}, as well as approaches that combine both perspectives into a unified account \citep{liu-strube-2025-joint}.

From a \textbf{narratology point of view}, texts maintain coherence not only through casual and temporal relations between events, but also through the sustained development of thematic situations and discourse focus as the story progresses \cite{Prince_1982}. Readers track how attention shifts between characters, events, and aspects of the story world while maintaining continuity with previously introduced elements, constructing a coherent mental representation of the narrative \citep{FludernikMonika2009Aitn}. 

These approaches capture complementary signals of coherence, and broadly correspond to the metrics we use, as they reflect how events, participants, and topics evolve over the course of a discourse. In multimodal narrative settings, coherence additionally depends on how the developing text is grounded in the visual context. Narratives must therefore maintain consistent references not only within the discourse itself but also with respect to the characters and events depicted in the accompanying image sequence \cite{Ryan-2001-narrative}.

In \textbf{visual storytelling}, evaluation has largely focused on grounding and surface-level text properties. GrooVIST \citep{surikuchi-etal-2023-groovist} and RoViST-VG \citep{wang-etal-2022-rovist} measure visual grounding by aligning extracted noun phrases with image content. Inter- and intra-story repetition \citep{yao-2019-intra-repetition} captures repetition rate using trigrams. \citet{yang-etal-2025-storyllava} combine these automatic metrics with human and LLM-based evaluation. In contrast to these approaches, our metrics target deeper aspects of narrative coherence.

\begin{table*}[t]
\centering
\setlength{\tabcolsep}{7pt} 
\begin{tabularx}{\linewidth}{@{}l l
    S[table-format=3.0]
    S[table-format=1.2]
    S[table-format=1.2]
    S[table-format=2.2]
    S[table-format=3.2]
    S[table-format=2.2]
    S[table-format=2.2]@{}}
\toprule
Prompt & System & {Seqs} & {Seg/Seq} & {Sent/Seg} & {Sent/Seq} & {Words/Seq} & {Words/Sent} & {Words/Seg} \\
\midrule
\multirow{2}{*}{Short} & Human & 60  & 5.80 & 1.29 & 7.48  & 76.48  & 10.22 & 13.19 \\
                           & LLMs  & 300 & 5.71 & 1.98 & 11.32 & 176.77 & 15.62 & 30.98 \\
\multirow{2}{*}{Long}    & Human & 180 & 5.70 & 2.12 & 12.07 & 164.91 & 13.66 & 28.93 \\
                           & LLMs  & 300 & 5.93 & 1.74 & 10.34 & 203.02 & 19.63 & 34.24 \\
\bottomrule
\end{tabularx}
\caption{General descriptive statistics of the data. Seqs = number of story sequences, Seg = segments, Sent = sentences, and Words = word tokens.
}
\label{tab:general_stats}
\end{table*}

\section{Experimental design}
\label{sec:data_design}

\begin{figure*}[!t]
\centering
\begin{minipage}{\textwidth}
\centering
\begin{minipage}[t]{0.18\textwidth}
\centering
\scriptsize\textbf{Image 1}\\
\vspace{0.2em}
\includegraphics[width=\linewidth]{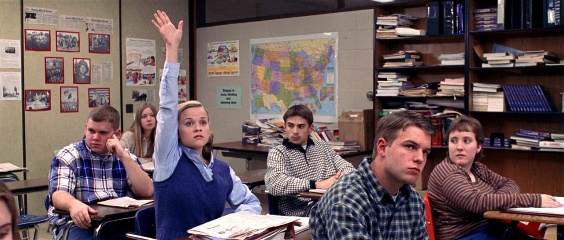}
\end{minipage}\hfill
\begin{minipage}[t]{0.18\textwidth}
\centering
\scriptsize\textbf{Image 2}\\
\vspace{0.2em}
\includegraphics[width=\linewidth]{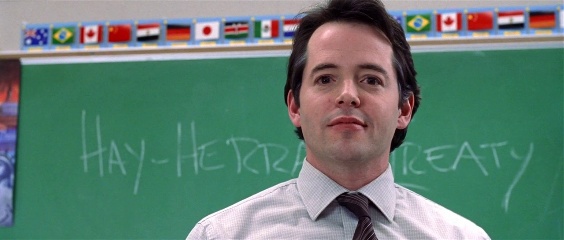}
\end{minipage}\hfill
\begin{minipage}[t]{0.18\textwidth}
\centering
\scriptsize\textbf{Image 3}\\
\vspace{0.2em}
\includegraphics[width=\linewidth]{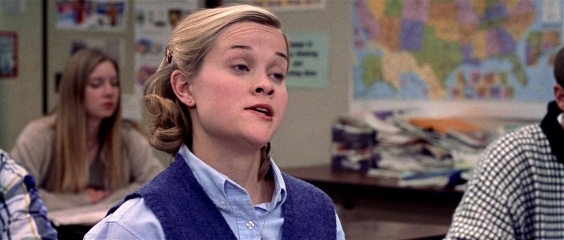}
\end{minipage}\hfill
\begin{minipage}[t]{0.18\textwidth}
\centering
\scriptsize\textbf{Image 4}\\
\vspace{0.2em}
\includegraphics[width=\linewidth]{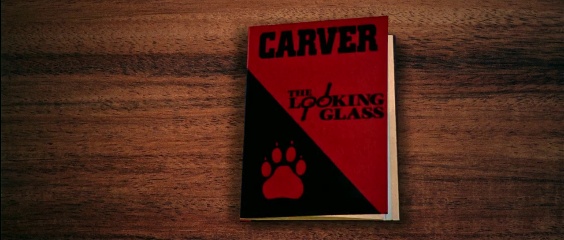}
\end{minipage}\hfill
\begin{minipage}[t]{0.18\textwidth}
\centering
\scriptsize\textbf{Image 5}\\
\vspace{0.2em}
\includegraphics[width=\linewidth]{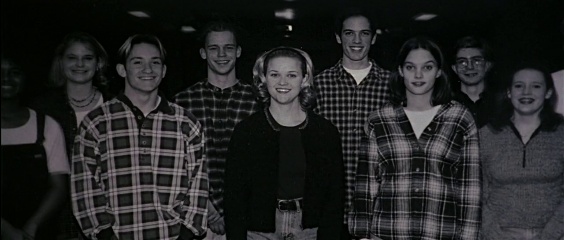}
\end{minipage}
\end{minipage}

\vspace{0.5em}

{\small
\RaggedRight
\sloppy
\setlength{\columnsep}{0.12in}
\setlength{\intextsep}{0pt}

\begin{wrapfigure}[12]{r}{0.95in}
\vspace{1.0em} 
\centering
\tiny\textbf{Reese}\\[-0.1em]
\vspace{1em}
\includegraphics[width=0.72in]{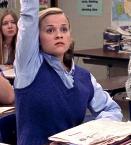}

\vspace{0.35em}

\tiny\textbf{Matthew}\\[-0.1em]
\vspace{1em}
\includegraphics[width=0.72in]{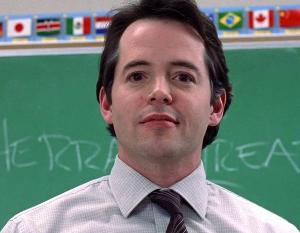}

\vspace{-0.6em}
\end{wrapfigure}

\noindent\rule{\linewidth}{0.2pt}
\colorbox{gray!12}{
\parbox{\dimexpr\linewidth-2\fboxsep}{
\textbf{\modelname{Human}, short:}
Reese is the only one in class who knows the answer to the teacher's question. \sepmark\ The teacher, Matthew notices that Reese is always first to answer and is very \ldots
}}

\noindent\rule{\linewidth}{0.2pt}
\colorbox{gray!12}{
\parbox{\dimexpr\linewidth-2\fboxsep}{
\textbf{\modelname{Human}, long:}
Reese was in it to win it. Win what? Didn't matter. As long as there were people around to beat, she was determined to come out on top. In this case, the setting was high school. History class. Senior year. \sepmark\ Matthew was a young teacher but had \ldots }}

\noindent\rule{\linewidth}{0.2pt}
\colorbox{gray!12}{
\parbox{\dimexpr\linewidth-2\fboxsep}{
\textbf{\modelname{GPT-4o}, short:}
In a lively classroom, Reese eagerly raises her hand, ready to answer a question. Her enthusiasm is evident among the other students, who seem less engaged. \sepmark\ At the front of the class, Matthew, the teacher, observes the students. He stands \ldots }} 

\noindent\rule{\linewidth}{0.2pt}
\colorbox{gray!12}{
\parbox{\dimexpr\linewidth-2\fboxsep}{
\textbf{\modelname{GPT-4o}, long:}
In a lively classroom, Reese eagerly raises her hand, surrounded by her attentive classmates. The room is filled with anticipation as they engage in a discussion, the walls adorned with educational posters and a map. \sepmark\ At the front of the class, \ldots }}

}
\caption{Example visual story sequence with numbered story and character images.
Four short excerpts are shown from human and GPT-4o outputs under the short and long prompt conditions. Both humans and models have access to the full sequence of images at once. The \sepmark\ marker indicates segment boundaries. Each segment is a chunk of text about a single image, in the same left-to-right order as the numbered images. Segments are used as the unit of analysis in several of our metrics. }
\label{fig:example_sequence}
\end{figure*}

\subsection{Data}
We use visual story sequences from the Visual Writing Prompts (VWP) corpus \citep{hong-etal-2023-visual-writing}, which contains human-written stories grounded in sequences of images. The visual sequence serves as the comparison unit in the paper. 

To determine the number of sequences needed for the analysis, we conducted an \textit{a priori} power analysis with G*Power 3.1 \citep{faul_statistical_2009}. Using standard settings (\(\alpha=.05\), power \(=.95\)) and assuming medium effect sizes, the analysis indicated that 46-75 visual sequences would be sufficient for our planned comparisons. Based on this estimate, we randomly sampled 60 visual story sequences from VWP.

\subsection{Models}

We generated stories for sampled story sequences using five multimodal language models: InternVL3-78B~\citep{zhu2025internvl3exploringadvancedtraining}, Qwen3-VL-235B~\citep{bai2025qwen3vltechnicalreport}, Claude-4.5 Sonnet\footnote{\url{https://www.anthropic.com/news/claude-sonnet-4-5}}, GPT-4o~\citep{openai2024gpt4technicalreport}, and Llama-4-Scout\footnote{\url{https://ai.meta.com/blog/llama-4-multimodal-intelligence/}}.
This set includes both open-source and proprietary models.
For each sequence, all models received the same visual input: the story image sequence together with character images and character names when available.

\subsection{Prompt conditions}
\label{subsec:prompts}

We evaluate narrative coherence under two prompt conditions that vary in the degree of guidance provided for story generation. The \textbf{short prompt} (\S \ref{sec:metrics_definitions}) is the primary condition and follows the original instructions from VWP's data collection \citep{hong-etal-2023-visual-writing}.
In this condition, human stories come from the VWP corpus, while model stories were newly generated. We modified the original instructions only by requiring models to insert \sepmark \, between image-level story segments.
%

The \textbf{long prompt} (\S \ref{sec:long_prompt}) provides more explicit guidance for story generation. In addition to the original task instructions, it emphasises tellability, coherence across the image sequence, and consistency in referring to characters. Under this condition, model stories were generated automatically, while additional human-written stories were collected through Amazon Mechanical Turk.

All analyses are based on normalised story texts. We standardised formatting and segment boundaries across all stories, and additionally removed reasoning traces, meta-commentary, and formatting artifacts from machine-generated texts. 

General descriptive statistics for human-/ and model-generated stories are reported in Table~\ref{tab:general_stats}.
All implementation details required to reproduce our experiments are provided in the GitHub repository accompanying this paper.


%
%

\section{Modelling narrative coherence: short prompt}
\label{sec:metrics_definitions}

To operationalise narrative coherence in visually grounded narratives, we define five metrics capturing complementary aspects of discourse organisation: coreference, implicit discourse relation typology, topic switch, character persistence, and multimodal character grounding. 

To compare overall coherence profiles, we combine these five metrics into a \textbf{narrative coherence score} (NCS), reported in arithmetic mean and geometric mean variants.
The purpose of this score is not to replace the individual metrics, but to provide a unified quantitative measure of the general pattern that they identify for a story.
The arithmetic mean reflects the unweighted average performance across metrics, while the geometric mean penalises imbalances among the scores more strongly. This distinction is useful because model-generated stories can do well on individual metrics, while still projecting uneven coherence profiles overall.

All metric scores are computed at the story level.
To place them on a comparable scale, we apply a $\tanh$ transformation before aggregation, similar to \citet{surikuchi-etal-2023-groovist}.
Unless otherwise noted, reported values are means across stories (standard deviations are parentheses).
For multi-component metrics, we also report the component values.
We additionally report paired $t$-tests comparing humans with each model.
Statistical significance is marked as $^{*}p<.05$, $^{**}p<.01$.
We always highlight human values in bold.
See Appendix~\ref{app:metric_details} for details on how the metrics were computed.

\subsection{Coreference}
We use coreference as a metric of referential continuity in the story, capturing how strongly reference is concentrated on recurring entities.
We converted all stories to a format compatible with the Link-Append coreference model \citep{10.1162/tacl_a_00543}. We used the implementation of \citet{porada-etal-2024-controlled-reevaluation} to extract coreference chains from human-/ and model-generated stories.
For a story with $C$ coreference chains and mean chain size $S$, we define the coreference score as $R = S/C$.
This score captures how consistently already introduced entities are re-mentioned over the course of the story narrative.

\begin{table}[!ht]
    \centering
    \resizebox{\linewidth}{!}{\begin{tabular}{lcccl}
    System & $C$ & $S$ & $R$ & $t$  \\
    \hline
    \modelname{Claude 4.5}    & 6.90 & 4.61 & 0.58 {\scriptsize(0.18)}  & 6.90$^{**}$ \\
    \modelname{GPT-4o}       & 5.88 & 4.73 & 0.65 {\scriptsize(0.21)}  & 4.60$^{**}$ \\
    \hline
    \modelname{InternVL3}    & 7.05 & 5.26 & 0.63 {\scriptsize(0.18)}  & 4.60$^{**}$ \\
    \modelname{Llama 4 Scout} & 4.87 & 5.43 & 0.73 {\scriptsize(0.23)}  & 1.12 \\
    \modelname{Qwen3-VL}     & 4.58 & 4.33 & 0.70 {\scriptsize(0.20)}  & 2.22$^{*}$ \\
    \hline
    \rowcolor{gray!12}\modelname{Humans}      & \textbf{3.80} & \textbf{4.67} & \textbf{0.77} {\scriptsize(0.19)}  & - \\
    \hline
    \end{tabular}}
    \caption{Coreference metric, short prompt.}
    \label{tab:coref-metrics-agg}
\end{table}

\paragraph{Results}
Table~\ref{tab:coref-metrics-agg} shows that humans achieve the highest coreference score ($R=0.77$), reflecting stronger referential continuity concentrated on a relatively small set of entities.
Humans also produce the fewest chains ($C=3.80$) and a comparatively large average chain size ($S=4.67$).
Models show a broader and less focused pattern of reference, with more chains and lower $R$ values.
Llama 4 Scout and Qwen3-VL are closest to the human score, while other models form a lower-scoring group.
Paired $t$-tests on story-level $R$ scores indicated statistically significant differences between humans and all models except Llama 4 Scout, with higher mean scores for humans.


\subsection{Implicit discourse relation typology}
We use implicit discourse relation typology as a measure of discourse variety across adjacent story segments in each narrative.

For each story, we classify the implicit discourse relation between neighbouring segments using DeDisCo \citep{ju-etal-2025-dedisco}\footnote{We use the publicly released checkpoint: \url{https://huggingface.co/JuNymphea/Georgetown-qwen3-4B-finetuned-for-disrpt2025}}, an instruction-tuned Qwen3-4B model \citep{yang2025qwen3technicalreport} released for the DISRPT 2025 shared task.
For a story with $U$ unique predicted relation types and $T$ total predicted relations, we define the typological diversity score as $D = U/T$.

\begin{table}[!ht]
    \centering
    \resizebox{\linewidth}{!}{\begin{tabular}{lcccl}
    System & $U$ & $T$ & $D$ & $t$  \\
    \hline
    \modelname{Claude 4.5}     & 2.13 & 4.67 & 0.43 {\scriptsize(0.13)} & 1.19 \\
    \modelname{GPT-4o}         & 2.02 & 4.60 & 0.41 {\scriptsize(0.15)} & 1.95 \\
    \hline
    \modelname{InternVL3}      & 1.83 & 4.95 & 0.35 {\scriptsize(0.15)} & 4.33$^{**}$ \\
    \modelname{Llama 4 Scout}  & 1.92 & 4.65 & 0.39 {\scriptsize(0.15)} & 2.63$^{*}$ \\
    \modelname{Qwen3-VL}       & 1.70 & 4.67 & 0.35 {\scriptsize(0.15)} & 4.50$^{**}$ \\
    \hline
    \rowcolor{gray!12}\modelname{Humans}         & \textbf{2.40} & \textbf{4.78} & \textbf{0.46} {\scriptsize(0.15)} & - \\
    \hline
    \end{tabular}}
    \caption{Implicit discourse relation typology, short prompt.}
    \label{tab:implicit-metrics-agg}
\end{table}

\paragraph{Results}
Table~\ref{tab:implicit-metrics-agg} shows that humans achieve the highest typological diversity score ($D=0.46$). 
Their texts also demonstrate the highest mean of unique relation types ($U=2.40$), while the total number of predicted relations ($T=4.78$) is close to the models' values for this part of the metric.
The difference therefore reflects typological diversity, rather than simple length.
Among models, Claude 4.5 and GPT-4o are closest to the human score, Llama 4 Scout occupies a middle position and InternVL-3 and Qwen3-VL form the lowest-scoring group.
Paired $t$-tests on story-level $D$ scores showed that humans scored significantly higher than open-source models, but not proprietary models.

\begin{figure}[!ht]
    \centering
    \includegraphics[width=.48\textwidth]{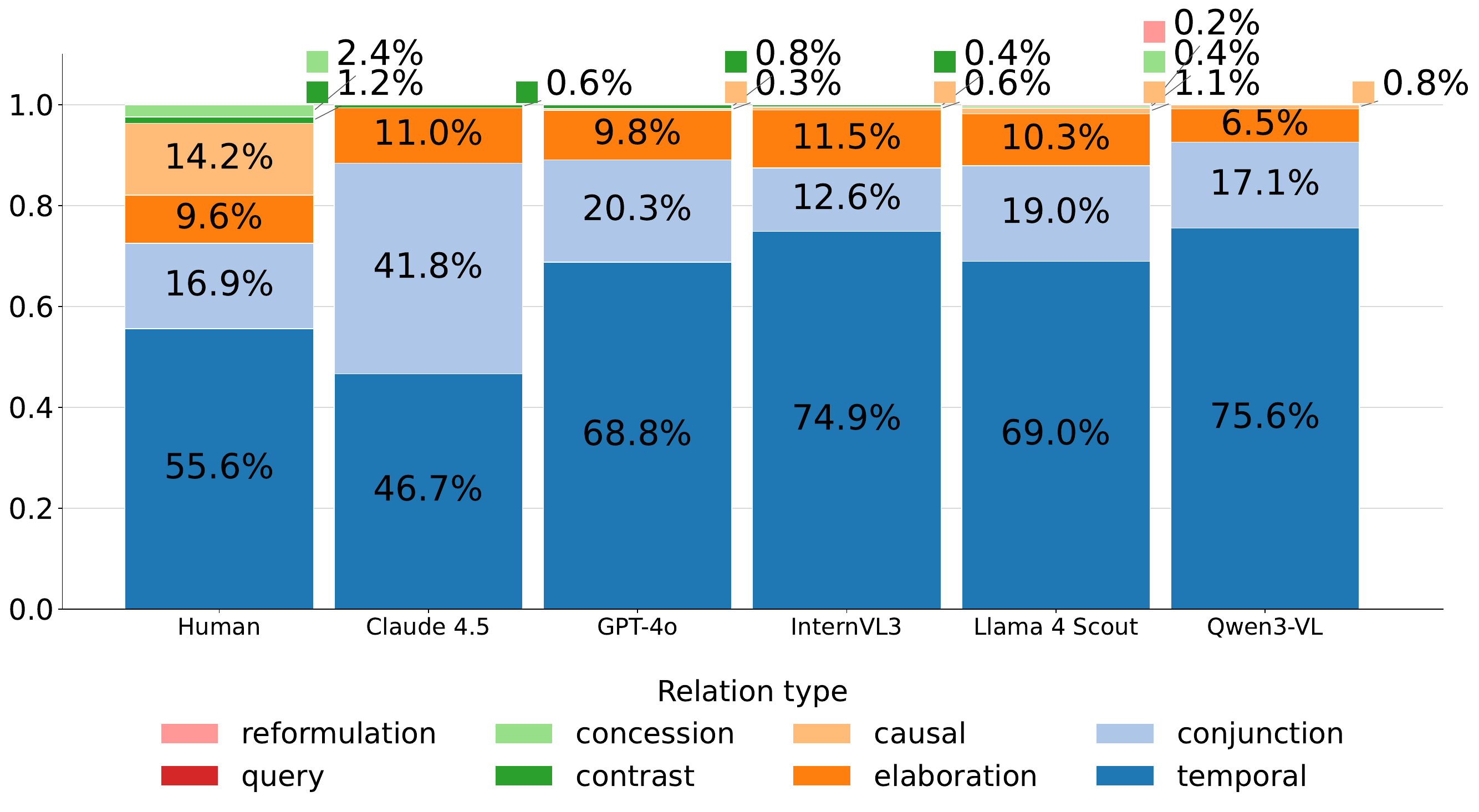}
    \caption{Implicit discourse relation type composition, short prompt.
    Each bar shows the mean within-story proportion of predicted implicit relation types, averaged across stories (and displayed as 100\% stacked bars). 
    }
    \label{fig:implicit_relation_heatmap}
\end{figure} 

Figure~\ref{fig:implicit_relation_heatmap} shows which relation types contribute to the observed pattern.
Human stories show a more mixed implicit relation profile with notably higher proportions of causal relations than any of the models.
Most models are more concentrated on temporal relations, and conjunction is also prominent for several systems, e.g., Claude 4.5.
Although Llama 4 Scout covers six relation types, matching humans in breadth, its distribution includes much less causal, and is more heavily dominated by temporal relations.

\subsection{Topic switch}
We use topic switching as a metric of topical progression. Specifically, we compute how often an assigned topic label changes between adjacent story segments. To obtain topic labels, we trained BERTopic \citep{grootendorst2022bertopic} on our corpus.
To reduce dependence on a single topic resolution, we compute topic switch under multiple topic granularities by applying BERTopic topic reduction over a range of topic counts using the $\mathbf{nr\_topics}$ parameter, and we report $T$ (topic switch) averaged across these settings.
%
To ensure comparability across prompt conditions, we train a single topic model on the combined corpus and keep it fixed for both short- and long-prompt robustness analyses (Section~\ref{sec:long_prompt}).
%
We additionally use balanced training to prevent the long-prompt human variants from dominating the learned topic space.
For a story with $N$ segments, the topic switch score $T$ is defined as the number of topic changes between adjacent segments divided by $N-1$.

\begin{table}[!ht]
    \centering
    \resizebox{\linewidth}{!}{\begin{tabular}{lccl}
    System & $N$ & $T$ & $t$ \\
    \hline
    \modelname{Claude 4.5}     & 5.67 & 0.27 {\scriptsize(0.20)} & 5.54$^{**}$ \\
    \modelname{GPT-4o}         & 5.60 & 0.28 {\scriptsize(0.18)} & 5.66$^{**}$ \\
    \hline
    \modelname{InternVL3}      & 5.95 & 0.27 {\scriptsize(0.18)} & 5.76$^{**}$ \\
    \modelname{Llama 4 Scout}  & 5.65 & 0.26 {\scriptsize(0.17)} & 5.76$^{**}$ \\
    \modelname{Qwen3-VL}       & 5.67 & 0.38 {\scriptsize(0.19)} & 1.65 \\
    \hline
    \rowcolor{gray!12}\modelname{Humans}         & \textbf{5.80} & \textbf{0.43} {\scriptsize(0.14)} & - \\
    \hline
    \end{tabular}}
    \caption{Topic switch, short prompt.}
    \label{tab:topic-switch-metrics-agg}
\end{table}

\paragraph{Results}
Table~\ref{tab:topic-switch-metrics-agg} shows that humans have the highest topic switch score ($T=0.43$), indicating more frequent topic transitions between adjacent segments than any of the models.
Qwen3-VL is closest to humans, while the remaining models form a lower, tightly clustered group.
Mean segment counts $N$ are very similar across systems, so these differences are unlikely to be driven by story length.
Paired $t$-tests on story-level $T$ scores showed that humans get significantly higher scores than all models except Qwen3-VL.

\begin{figure}[!ht]
    \centering
    \includegraphics[width=.48\textwidth]{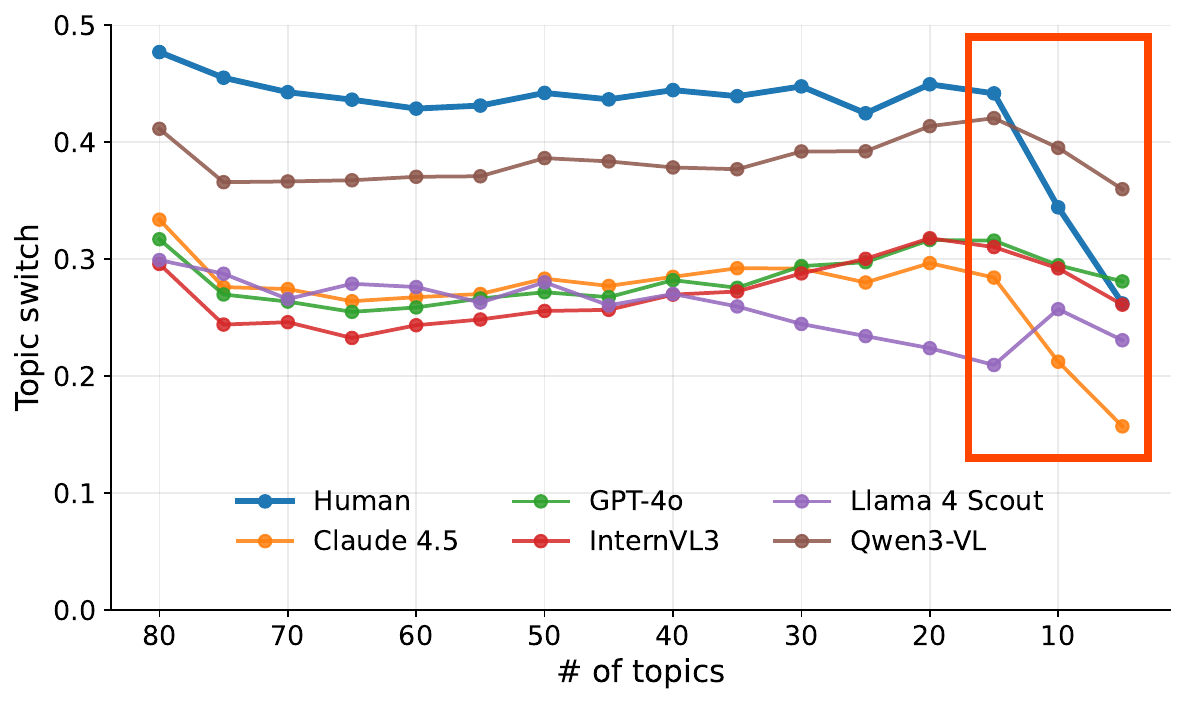}
    \caption{Topic switch under topic space compression.
    The highlighted region ($\mathbf{nr\_topics}=15$ to $5$) marks a pronounced drop in human topic switch.
    }
    \label{fig:topic_switch_rate}
\end{figure}

\paragraph{Effect of reducing topic space}
We vary topic resolution by decreasing $\mathbf{nr\_topics}$ from 80 to 5 in increments of 5, tracking topic switch at each step. Lower $\mathbf{nr\_topics}$ enforces stronger topic merging. This increases the likelihood that adjacent segments share the same label, and so the switch rate is expected to decline. Indeed, as Figure \ref{fig:topic_switch_rate} shows, all systems show lower switch rates as the topic space is compressed. The human curve remains highest for most settings, indicating greater topical variation when more finer-grained topics are available. The highlighted region in Figure \ref{fig:topic_switch_rate} shows a clear shift: the human curve drops more sharply than most model curves. This suggests that human topic variation depends more strongly on fine-grained topic distinctions, while model topic structure is already coarse and therefore less affected by forced topic compression.

\subsection{Character persistence}
Character persistence measures character continuity across the story. Using MovieNet character annotations \citep{huang2020movienetholisticdatasetmovie} linked to VWP stories, we first extract the set of characters associated with each story's image sequence.
To locate these characters in text, we run the Link-Append coreference model and identify coreference chains whose mentions match the annotated character names.
From these aligned character mentions, we compute two complementary sub-metrics.
\texttt{ChC} (\textbf{Ch}aracter \textbf{C}ontinuity) captures how consistently a character is mentioned across neighbouring story segments.
\texttt{ChS} (\textbf{Ch}aracter \textbf{S}pread) captures how broadly the character is distributed across the full story.
We then combine these into a single \textbf{Ch}aracter \textbf{P}ersistence score, \texttt{ChP}.
If a story has no valid character annotation or no character-coreference match, we assign a score of 0 for that case.

\begin{table}[!ht]
    \centering
    \resizebox{\linewidth}{!}{
    \begin{tabular}{lcccl}
    System & ChC & ChS & ChP & $t$ \\
    \hline
    \modelname{Claude 4.5}     & 0.27 & 0.47 & 0.32 {\scriptsize(0.29)} & 1.93 \\
    \modelname{GPT-4o}         & 0.35 & 0.51 & 0.36 {\scriptsize(0.29)} & 1.02 \\
    \hline
    \modelname{InternVL3}      & 0.43 & 0.60 & 0.38 {\scriptsize(0.31)} & 0.35 \\
    \modelname{Llama 4 Scout}  & 0.07 & 0.10 & 0.08 {\scriptsize(0.22)} & 7.18$^{**}$ \\
    \modelname{Qwen3-VL}       & 0.28 & 0.52 & 0.30 {\scriptsize(0.26)} & 2.50$^{*}$ \\
    \hline
    \rowcolor{gray!12}\modelname{Humans}         & \textbf{0.36} & \textbf{0.52} & \textbf{0.39} {\scriptsize(0.29)} & - \\
    \hline
    \end{tabular}
    }
    \caption{Character metrics, short prompt.}
    \label{tab:character-metrics-agg}
\end{table}

\paragraph{Results}
Table~\ref{tab:character-metrics-agg} shows that humans achieve the highest character persistence score ($0.39$), indicating stronger continuity of story characters over time than any of the models.
InternVL-3 and GPT-4o are closest to the human score, while Claude 4.5 and Qwen3-VL score somewhat lower.
Llama 4 Scout is substantially lower than all other systems.
Interestingly, InternVL-3 has the highest values for both \texttt{ChC} and \texttt{ChS}, but not the highest persistence score.
Paired $t$-tests on story-level \texttt{ChP} scores showed that humans get significantly higher scores than Llama 4 Scout and Qwen3-VL.

\subsection{Multimodal character grounding}

We define multimodal character grounding as a measure of multimodal coherence specific to story characters.
Unlike character persistence, this metric explicitly incorporates visual modality.
The metric consists of two components.
First, multimodal character continuity (MCC) is computed. It measures how closely continuous reference to characters in text matches appearances of characters in the visual sequence.
Second, GROOViST (GV) \citep{surikuchi-etal-2023-groovist} provides a complementary story-level measure of visual grounding.
We combine these components into a single multimodal character grounding (MCG) score.

\begin{table}[!ht]
    \centering
    \resizebox{\linewidth}{!}{\begin{tabular}{lcccl}
    System & GV & MCC & MCG & $t$ \\
    \hline
    \modelname{Claude 4.5}     & 0.75 & 0.41 & 0.46 {\scriptsize(0.28)} & -0.20 \\
    \modelname{GPT-4o}         & 0.56 & 0.45 & 0.58 {\scriptsize(0.34)} & -2.13$^{*}$ \\
    \hline
    \modelname{InternVL3}      & 0.60 & 0.41 & 0.53 {\scriptsize(0.34)} & -1.35 \\
    \modelname{Llama 4 Scout}  & 0.35 & 0.10 & 0.15 {\scriptsize(0.32)} & 4.46$^{**}$ \\
    \modelname{Qwen3-VL}       & 0.72 & 0.45 & 0.50 {\scriptsize(0.30)} & -0.78 \\
    \hline
    \rowcolor{gray!12}\modelname{Humans}         & \textbf{0.45} & \textbf{0.38} & \textbf{0.45} {\scriptsize(0.49)} & - \\
    \hline
    \end{tabular}}
    \caption{Multimodal character grounding, short prompt.}
    \label{tab:cgc-metrics-agg}
\end{table}

\paragraph{Results}
Table~\ref{tab:cgc-metrics-agg} differs from the earlier metrics in that it does not show a clear human-leading pattern.
Humans fall near the middle of the system range on GV, MCC, and MCG, while some models achieve higher MCG values, especially GPT-4o.
This indicates that some models produce narratives that are tightly linked to visually present characters.
Human stories are less visually literal as indicated by lower scores compared to the models, moving beyond what is directly depicted, rather than re-mentioning characters simply because they remain visible.
Llama 4 Scout is a clear outlier: it scores low on all three metrics, consistent with earlier evidence that its textual continuity is only weakly tied to the expected image-grounded characters.
Paired $t$-tests on story-level MCG showed that humans score significantly lower than GPT-4o and significantly higher than Llama 4 Scout.

\begin{table*}[!ht]
    \centering
    \begin{tabular}{lclcl}
    System & $\mathrm{NCS}^{\text{arith}}$ & $t_{\text{arith}}$ & $\mathrm{NCS}^{\text{geom}}$ & $t_{\text{geom}}$ \\
    \hline
    \modelname{Claude 4.5}     & 0.41 {\scriptsize(0.12)} & 4.68$^{**}$ & 0.27 {\scriptsize(0.21)} & 2.84$^{**}$ \\
    \modelname{GPT-4o}         & 0.46 {\scriptsize(0.12)} & 2.58$^{*}$ & 0.32 {\scriptsize(0.21)} & 1.77 \\
    \hline
    \modelname{InternVL3}      & 0.43 {\scriptsize(0.13)} & 3.72$^{**}$ & 0.30 {\scriptsize(0.20)} & 2.01$^{*}$ \\
    \modelname{Llama 4 Scout}  & 0.32 {\scriptsize(0.10)} & 8.14$^{**}$ & 0.06 {\scriptsize(0.15)} & 8.21$^{**}$ \\
    \modelname{Qwen3-VL}       & 0.45 {\scriptsize(0.11)} & 2.77$^{**}$ & 0.32 {\scriptsize(0.22)} & 1.59 \\
    \hline
    \rowcolor{gray!12}\modelname{Humans}         & \textbf{0.50} {\scriptsize(0.14)} & - & \textbf{0.36} {\scriptsize(0.27)} & - \\
    \hline
    \end{tabular}
    \caption{Narrative coherence scores, short prompt.
    $t_{\text{arith}}$ and $t_{\text{geom}}$ are human-vs-model paired tests. 
    }
    \label{tab:ncs-metrics-agg-60}
\end{table*}

\subsection{Narrative coherence score}
We define the narrative coherence score (NCS) as a composite story-level measure that combines the five coherence metrics into a single value capturing coherence across multiple dimensions. For each story, we compute two variants by aggregating the five metric values. The arithmetic variant, $\mathrm{NCS}^{\text{arith}}$, is the mean of the component scores, while the geometric variant, $\mathrm{NCS}^{\text{geom}}$, is more sensitive to imbalance across dimensions and therefore penalises stories that are strong on some metrics but weak on others. Higher values on both variants indicate greater coherence under the dimensions captured by our framework.

\paragraph{Results}
As shown in Table~\ref{tab:ncs-metrics-agg-60}, humans achieve the highest score on both NCS variants.
For $\mathrm{NCS}^{\text{arith}}$, humans score $0.50$, with most models forming a relatively tight cluster below them, from $0.41$ to $0.46$.
Llama 4 Scout stands apart as a substantially lower-scoring outlier.
The same overall pattern holds for $\mathrm{NCS}^{\text{geom}}$, but the differences between models are larger.
Humans again score the highest ($0.36$).
Llama 4 Scout drops to $0.06$ indicating a particularly uneven coherence profile across individual metrics.
On the arithmetic variant, paired $t$-tests showed that humans scored significantly higher than every model.
On the geometric variant, humans score significantly higher than Claude 4.5 and Llama 4 Scout.

\subsection{Summary of the short prompt results}
\label{sec:summary_short}

The short-prompt results suggest that human- and model-generated stories differ more in their overall coherence patterns than in any single metric on its own.
%
In text-oriented metrics, human stories show a consistent profile: tighter referential focus, greater discourse relation diversity, more frequent topic shifts, and stronger character persistence. In contrast, multimodal character grounding shows a different pattern, with some models matching or exceeding human values. This suggests that stronger visual grounding does not mean better storytelling in the human sense. Human stories are less constrained by what is directly visible, while current models may still place considerable weight on staying close to the image content.
%
%
The difference lies less in any single metric than in the consistent pattern across them. The metrics capture different aspects of narrative organisation and do not all favour the same source. Under the short prompt, the NCS makes the human–model difference clearer than any individual metric. Section~\ref{sec:long_prompt} examines whether this pattern changes under more explicit guidance by modifying both the human instructions and the prompts for the models.

\section{Robustness analysis: long prompt}
\label{sec:long_prompt}

\paragraph{Data collection}
For the long-prompt condition, we collected new human-written narratives through Amazon Mechanical Turk (AMT), while model outputs were generated using the same prompt.
We obtained three human descriptions for each of 60 visual story sequences.
Workers received a base payment of $\$4.00$ per story together with word-based bonuses, for a total data collection cost of $\$2,303.62$ including platform fees.

\paragraph{Quality control}
We applied an additional quality control check to the newly collected human long-prompt data to identify cases that were potentially AI-assisted or AI-generated.
We fine-tuned a RoBERTa-based binary text classifier \citep{liu_roberta_2019} to distinguish long-prompt human stories from long-prompt model-generated stories in our dataset, and used its outputs as a signal for story exclusion.
We used 5-fold cross-validation grouped by visual sequence: folds were split so that all stories associated with a given sequence appeared in the same fold.
This means that when a story was scored, the classifier had not seen that story or any other story paired with the same visual sequence during training.
Because comparisons are made at the visual sequence level, the exclusion rule was conservative: if at least one human story for a sequence was flagged with AI-generated probability greater than $0.9$, the entire sequence was removed from the long-prompt analysis.
Under this rule, 54 of the 60 visual sequences were kept.
All long-prompt results described next are therefore based on these 54 sequences.

\begin{table*}[!ht]
    \centering
\begin{tabular}{
    >{\arraybackslash}p{3cm} |
    >{\centering\arraybackslash}p{2cm}
    >{\centering\arraybackslash}p{2cm} |
    >{\centering\arraybackslash}p{1.5cm}
    >{\centering\arraybackslash}p{1.5cm}
    >{\centering\arraybackslash}p{1.5cm}
}    System
      & Short & Long & $\Delta_{\text{short}}$ & $\Delta_{\text{long}}$ & $t$ \\
    \hline
    \modelname{Claude 4.5}
      & 0.42 {\scriptsize(0.12)} & 0.41 {\scriptsize(0.12)} & 0.083 & 0.068 & 0.86 \\
    \modelname{GPT-4o}
      & 0.45 {\scriptsize(0.11)} & 0.46 {\scriptsize(0.11)} & 0.040 & 0.012 & 1.73 \\
    \hline
    \modelname{InternVL3}
      & 0.43 {\scriptsize(0.12)} & 0.40 {\scriptsize(0.13)} & 0.068 & 0.074 & -0.31 \\
    \modelname{Llama 4 Scout}
      & 0.32 {\scriptsize(0.10)} & 0.38 {\scriptsize(0.14)} & 0.182 & 0.095 & 2.77$^{**}$ \\
    \modelname{Qwen3-VL}
      & 0.45 {\scriptsize(0.11)} & 0.41 {\scriptsize(0.11)} & 0.052 & 0.065 & -0.63 \\
    \hline
    \rowcolor{gray!12}\modelname{Humans}
      & \textbf{0.50} {\scriptsize(0.14)} & \textbf{0.48} {\scriptsize(0.11)} & -- & -- & -- \\
    \hline
    \end{tabular}
    \\[0.5em]
    {\small (a) $\mathrm{NCS}^{\text{arith}}$}
    \\[0.5em]
\begin{tabular}{
    >{\arraybackslash}p{3cm} |
    >{\centering\arraybackslash}p{2cm}
    >{\centering\arraybackslash}p{2cm} |
    >{\centering\arraybackslash}p{1.5cm}
    >{\centering\arraybackslash}p{1.5cm}
    >{\centering\arraybackslash}p{1.5cm}
}    
    System
      & Short & Long & $\Delta_{\text{short}}$ & $\Delta_{\text{long}}$ & $t$ \\
    \hline
    \modelname{Claude 4.5}
      & 0.29 {\scriptsize(0.21)} & 0.26 {\scriptsize(0.21)} & 0.085 & 0.112 & -0.85 \\
    \modelname{GPT-4o}
      & 0.32 {\scriptsize(0.20)} & 0.33 {\scriptsize(0.21)} & 0.055 & 0.040 & 0.54 \\
    \hline
    \modelname{InternVL3}
      & 0.30 {\scriptsize(0.20)} & 0.26 {\scriptsize(0.20)} & 0.069 & 0.110 & -1.38 \\
    \modelname{Llama 4 Scout}
      & 0.06 {\scriptsize(0.15)} & 0.17 {\scriptsize(0.24)} & 0.314 & 0.204 & 2.44$^{*}$ \\
    \modelname{Qwen3-VL}
      & 0.32 {\scriptsize(0.21)} & 0.27 {\scriptsize(0.20)} & 0.049 & 0.104 & -1.67 \\
    \hline
    \rowcolor{gray!12}\modelname{Humans}
      & \textbf{0.37} {\scriptsize(0.26)} & \textbf{0.37} {\scriptsize(0.20)} & -- & -- & -- \\
    \hline
    \end{tabular}
    \\[0.5em]
    {\small (b) $\mathrm{NCS}^{\text{geom}}$}

    \caption{Narrative coherence scores on the same 54-story subset under short and long prompts. Table~(a) reports results for $\mathrm{NCS}^{\text{arith}}$, and Table~(b) reports results for $\mathrm{NCS}^{\text{geom}}$. $\Delta_{\text{short}}$ and $\Delta_{\text{long}}$ are the human-model gaps under short and long prompts, respectively, and $t$ is the paired test statistic comparing the two gaps. $^{*}p<.05$, $^{**}p<.01$.}
    \label{tab:ncs-metrics-gap-change-54}
\end{table*}

\paragraph{Results}
Table~\ref{tab:ncs-metrics-gap-change-54} reports NCS values for both prompt conditions on the same subset of 54 sequences, together with paired tests of whether the human-model gap changed across prompts.
Under the long prompt, humans remain the highest-scoring source on both variants of NCS. The overall human-model separation observed under the short prompt is mostly preserved.
Prompt effects are model-specific: only Llama 4 Scout shows a significant reduction in the human-model gap.

At the level of individual metrics, the long prompt produces selective rather than uniform changes. One point of convergence appears in topic switch: human scores decrease from $0.42$ to $0.34$, while GPT-4o increases from $0.27$ to $0.31$ and Intern-VL3 from $0.27$ to $0.29$. LLama 4 Scout slightly exceeds the human value, increasing from $0.25$ to $0.36$. Implicit discourse relation diversity shows a similar, though weaker, pattern for some systems: the human score decreases from $0.45$ to $0.41$, while Qwen3-VL increases from $0.34$ to $0.40$. Coreference changes less, with the human score dropping from $0.79$ to $0.69$ and only small gains for most models (e.g., GPT-4o from $0.64$ to $0.66$). The largest improvements in character persistence and multimodal character grounding appear in Llama 4 Scout's outputs (\texttt{ChP}: $0.08$ to $0.20$; MCG: $0.15$ to $0.36$), while other models show mixed or negative changes on these metrics.

Taken together, these results suggest that the long prompt condition does not produce a general convergence toward the human coherence profile, but benefits some systems more than others.

\begin{table*}[!ht]
\centering
\setlength{\tabcolsep}{5pt}
\begin{tabular}{lccccc}
& \multicolumn{1}{c}{Scraped data} & \multicolumn{2}{c}{Short prompt} & \multicolumn{2}{c}{Long prompt} \\
\cmidrule(lr){2-2}\cmidrule(lr){3-4}\cmidrule(lr){5-6}
Evaluator & Human & Human & Models & Human & Models \\
\midrule
\modelname{qwen3vl}
& \cellcolor{rHigh} 14.21 {\scriptsize(8.32)}
& \cellcolor{rHigh} 13.58 {\scriptsize(5.21)}
& \cellcolor{rLow} 3.12--4.31
& \cellcolor{rMid} 11.54 {\scriptsize(2.43)}
& \cellcolor{rLow} 2.67--4.11 \\

\modelname{llama4scout}
& \cellcolor{rTop}\textcolor{white}{37.00 {\scriptsize(36.44)}}
& \cellcolor{rTop}\textcolor{white}{25.98 {\scriptsize(11.69)}}
& \cellcolor{rMid} 1.83--18.04
& \cellcolor{rHigh} 20.85 {\scriptsize(5.34)}
& \cellcolor{rMid} 1.67--14.42 \\

\modelname{internvl3}
& \cellcolor{rMid} 9.39 {\scriptsize(3.19)}
& \cellcolor{rMid} 11.44 {\scriptsize(4.48)}
& \cellcolor{rLow} 2.41--5.93
& \cellcolor{rMid} 10.30 {\scriptsize(2.06)}
& \cellcolor{rLow} 3.39--5.05 \\
\bottomrule
\end{tabular}
\caption{Perplexity by evaluator.
Columns are grouped by data condition.
Human cells report mean perplexity for human-authored texts, with standard deviations in parentheses.
Model cells report the range of mean perplexities across model-generated sources under the same evaluator.
Darker shading indicates higher perplexity.}
\label{tab:perplexity_means}
\end{table*}

\section{Discussion}

Our results show that human- and model-generated visual stories differ from one another in terms of their coherence profile.
Human stories score higher on the text-oriented measures (coreference, implicit discourse relation diversity, topic switch, and character persistence) while some models match or exceed humans on multimodal character grounding.
This is important because our framework is not meant to show that humans dominate every metric, but that humans and models organise narrative information differently across these metrics.
This also suggests that stronger visual grounding does not, in itself, imply more human-like narrative coherence.

The contrast is well captured by the composite narrative coherence score. Although model scores are sometimes close to human values in several individual metrics, the combined metric shows a more consistent human-model separation.

The long-prompt results further show that coherence is only partially subject to influence through corrective prompting.
More explicit instructions lead to selective improvements for some systems, most clearly Llama 4 Scout, but they do not remove the broader difference between human and model coherence profiles.

\paragraph{Perplexity as complementary evidence}
Our metrics show that human-generated stories differ from model-generated texts in their discourse organisation. To test whether this difference appears beyond the coherence measures in our framework, we examine perplexity as a complementary probe. Unlike our metrics, perplexity does not target a particular aspect of coherence directly. Instead, it measures how well a model predicts a given text, allowing us to assess whether human and model-generated stories differ in how probable they are under the models. This is relevant in our setting because several models occupy similar ranges on individual coherence metrics, even when they remain distinct from humans overall.


We compute perplexity with three open-source VLMs as evaluator models: Qwen3-VL, Llama 4 Scout, Intern-VL3.
We evaluate models on data from both prompt conditions and on additional multimodal data scraped from the web.
In all settings, perplexity is computed over text tokens.

Results, displayed in Table~\ref{tab:perplexity_means}, show a consistent pattern across evaluators and conditions.
All three evaluator models assign  higher perplexity to human-authored texts than to model-generated texts.
Although human perplexity is somewhat lower under the long prompt, it remains above the model range for every evaluator.
This suggests that human-authored narratives are less predictable under the open-source models that we evaluated than model-generated ones. In information theoretic terms, the human stories have higher surprisal values than the model generated narratives. This is also in line with additional findings reporting that LLM output tend to be homogeneous, both with and across models, in contrast to the diversity of human writing \citep{jiang2025-artificial-hivemind}.

We note that published stories from the VWP dataset may have been seen during pre-training by some evaluator models.
Thus, the VWP results do not constitute a strict test on unseen data.
However, such prior exposure would be expected to lower perplexity.
The fact that human-authored stories still receive higher perplexity than model-generated ones therefore makes the observed pattern more striking.

\section{Conclusion}

We introduced a framework for analysing narrative coherence in visually grounded storytelling, and we used it to compare human-written and model-generated stories from the VWP dataset \citep{hong-etal-2023-visual-writing}.
Across our coherence metrics, human and model stories showed systematically different profiles.
While individual differences were sometimes subtle, a consistent pattern emerged through an aggregated narrative coherence score.
We also found that explicit guidance for models to generate more coherent stories produces only limited convergence toward the human profile. 
Our results indicate that while current VLMs generate fluent, visually grounded narratives, they structure them in ways that significantly diverge from human discourse about visual scenes.



\bibliography{references}

@inproceedings{liu-strube-2025-joint,
    title = "Joint Modeling of Entities and Discourse Relations for Coherence Assessment",
    author = "Liu, Wei  and
      Strube, Michael",
    editor = "Christodoulopoulos, Christos  and
      Chakraborty, Tanmoy  and
      Rose, Carolyn  and
      Peng, Violet",
    booktitle = "Proceedings of the 2025 Conference on Empirical Methods in Natural Language Processing",
    month = nov,
    year = "2025",
    address = "Suzhou, China",
    publisher = "Association for Computational Linguistics",
    url = "https://aclanthology.org/2025.emnlp-main.1113/",
    doi = "10.18653/v1/2025.emnlp-main.1113",
    pages = "21910--21926",
    ISBN = "979-8-89176-332-6"
}

@book{Ryan-2001-narrative,
author = {Ryan, Marie-Laure},
title = {Narrative as Virtual Reality: Immersion and Interactivity in Literature and Electronic Media},
year = {2001},
isbn = {0801864887},
publisher = {Johns Hopkins University Press},
address = {USA}
}

@book{FludernikMonika2009Aitn,
publisher = {Routledge},
isbn = {9780203882887},
year = {2009},
title = {An introduction to narratology},
language = {eng},
address = {London ; New York},
author = {Fludernik, Monika},
}

@book{Prince_1982,
url = {https://doi.org/10.1515/9783110838626},
title = {Narratology: The Form and Functioning of Narrative},
author = {Gerald Prince},
publisher = {De Gruyter Mouton},
series = {Janua linguarum. Series Maior 108},
address = {Berlin},
doi = {doi:10.1515/9783110838626},
isbn = {9783110838626},
year = {1982},
lastchecked = {2026-03-12}
}

@inproceedings{rohde-etal-2018-discourse,
    title = "Discourse Coherence: Concurrent Explicit and Implicit Relations",
    author = "Rohde, Hannah  and
      Johnson, Alexander  and
      Schneider, Nathan  and
      Webber, Bonnie",
    editor = "Gurevych, Iryna  and
      Miyao, Yusuke",
    booktitle = "Proceedings of the 56th Annual Meeting of the Association for Computational Linguistics (Volume 1: Long Papers)",
    month = jul,
    year = "2018",
    address = "Melbourne, Australia",
    publisher = "Association for Computational Linguistics",
    url = "https://aclanthology.org/P18-1210/",
    doi = "10.18653/v1/P18-1210",
    pages = "2257--2267"
}

@inproceedings{lin-etal-2011-automatically,
    title = "Automatically Evaluating Text Coherence Using Discourse Relations",
    author = "Lin, Ziheng  and
      Ng, Hwee Tou  and
      Kan, Min-Yen",
    editor = "Lin, Dekang  and
      Matsumoto, Yuji  and
      Mihalcea, Rada",
    booktitle = "Proceedings of the 49th Annual Meeting of the Association for Computational Linguistics: Human Language Technologies",
    month = jun,
    year = "2011",
    address = "Portland, Oregon, USA",
    publisher = "Association for Computational Linguistics",
    url = "https://aclanthology.org/P11-1100/",
    pages = "997--1006"
}

@inproceedings{elsner-charniak-2011-extending,
    title = "Extending the Entity Grid with Entity-Specific Features",
    author = "Elsner, Micha  and
      Charniak, Eugene",
    editor = "Lin, Dekang  and
      Matsumoto, Yuji  and
      Mihalcea, Rada",
    booktitle = "Proceedings of the 49th Annual Meeting of the Association for Computational Linguistics: Human Language Technologies",
    month = jun,
    year = "2011",
    address = "Portland, Oregon, USA",
    publisher = "Association for Computational Linguistics",
    url = "https://aclanthology.org/P11-2022/",
    pages = "125--129"
}

@article{barzilay-lapata-2008-modeling,
    title = "Modeling Local Coherence: An Entity-Based Approach",
    author = "Barzilay, Regina  and
      Lapata, Mirella",
    journal = "Computational Linguistics",
    volume = "34",
    number = "1",
    year = "2008",
    url = "https://aclanthology.org/J08-1001/",
    doi = "10.1162/coli.2008.34.1.1",
    pages = "1--34"
}

@InProceedings{namuduri-qudsim_2025,
  author =  {Ramya Namuduri and Yating Wu and Anshun Asher Zheng and Manya Wadhwa and Greg Durrett and Junyi Jessy Li} ,
  title =   "QUDsim: Quantifying Discourse Similarities in LLM-Generated Text",
  booktitle =   "Proceedings of the Conference on Language Modeling (COLM)",
  year =    "2025",
  url =     "https://arxiv.org/abs/2504.09373"
}

@misc{huang2020movienetholisticdatasetmovie,
      title={MovieNet: A Holistic Dataset for Movie Understanding}, 
      author={Qingqiu Huang and Yu Xiong and Anyi Rao and Jiaze Wang and Dahua Lin},
      year={2020},
      eprint={2007.10937},
      archivePrefix={arXiv},
      primaryClass={cs.CV},
      url={https://arxiv.org/abs/2007.10937}, 
}

@article{liu_roberta_2019,
  author       = {Yinhan Liu and
                  Myle Ott and
                  Naman Goyal and
                  Jingfei Du and
                  Mandar Joshi and
                  Danqi Chen and
                  Omer Levy and
                  Mike Lewis and
                  Luke Zettlemoyer and
                  Veselin Stoyanov},
  title        = {RoBERTa: {A} Robustly Optimized {BERT} Pretraining Approach},
  journal      = {CoRR},
  volume       = {abs/1907.11692},
  year         = {2019},
  url          = {http://arxiv.org/abs/1907.11692},
  eprinttype    = {arXiv},
  eprint       = {1907.11692},
  bibsource    = {dblp computer science bibliography, https://dblp.org}
}

@inproceedings{ju-etal-2025-dedisco,
    title = "{D}e{D}is{C}o at the {DISRPT} 2025 Shared Task: A System for Discourse Relation Classification",
    author = "Ju, Zhuoxuan  and
      Wu, Jingni  and
      Purushothama, Abhishek  and
      Zeldes, Amir",
    editor = "Braud, Chlo{\'e}  and
      Liu, Yang Janet  and
      Muller, Philippe  and
      Zeldes, Amir  and
      Li, Chuyuan",
    booktitle = "Proceedings of the 4th Shared Task on Discourse Relation Parsing and Treebanking (DISRPT 2025)",
    month = nov,
    year = "2025",
    address = "Suzhou, China",
    publisher = "Association for Computational Linguistics",
    url = "https://aclanthology.org/2025.disrpt-1.4/",
    doi = "10.18653/v1/2025.disrpt-1.4",
    pages = "48--62",
    ISBN = "979-8-89176-344-9"
}

@article{faul_statistical_2009,
	title = {Statistical power analyses using {G}*{Power} 3.1: {Tests} for correlation and regression analyses},
	volume = {41},
	issn = {1554-3528},
	url = {https://doi.org/10.3758/BRM.41.4.1149},
	doi = {10.3758/BRM.41.4.1149},
	number = {4},
	journal = {Behavior Research Methods},
	author = {Faul, Franz and Erdfelder, Edgar and Buchner, Axel and Lang, Albert-Georg},
	month = nov,
	year = {2009},
	pages = {1149--1160},
}

@inproceedings{yao-2019-intra-repetition,
author = {Yao, Lili and Peng, Nanyun and Weischedel, Ralph and Knight, Kevin and Zhao, Dongyan and Yan, Rui},
title = {Plan-and-write: towards better automatic storytelling},
year = {2019},
isbn = {978-1-57735-809-1},
publisher = {AAAI Press},
url = {https://doi.org/10.1609/aaai.v33i01.33017378},
doi = {10.1609/aaai.v33i01.33017378},
booktitle = {Proceedings of the Thirty-Third AAAI Conference on Artificial Intelligence and Thirty-First Innovative Applications of Artificial Intelligence Conference and Ninth AAAI Symposium on Educational Advances in Artificial Intelligence},
articleno = {906},
numpages = {8},
location = {Honolulu, Hawaii, USA},
series = {AAAI'19/IAAI'19/EAAI'19}
}

@inproceedings{wang-etal-2022-rovist,
    title = "{R}o{V}i{ST}: Learning Robust Metrics for Visual Storytelling",
    author = "Wang, Eileen  and
      Han, Caren  and
      Poon, Josiah",
    editor = "Carpuat, Marine  and
      de Marneffe, Marie-Catherine  and
      Meza Ruiz, Ivan Vladimir",
    booktitle = "Findings of the Association for Computational Linguistics: NAACL 2022",
    month = jul,
    year = "2022",
    address = "Seattle, United States",
    publisher = "Association for Computational Linguistics",
    url = "https://aclanthology.org/2022.findings-naacl.206/",
    doi = "10.18653/v1/2022.findings-naacl.206",
    pages = "2691--2702",
}

@inproceedings{yang-etal-2025-storyllava,
    title = "{S}tory{LL}a{VA}: Enhancing Visual Storytelling with Multi-Modal Large Language Models",
    author = "Yang, Li  and
      Xiao, Zhiding  and
      Huang, Wenxin  and
      Zhong, Xian",
    editor = "Rambow, Owen  and
      Wanner, Leo  and
      Apidianaki, Marianna  and
      Al-Khalifa, Hend  and
      Eugenio, Barbara Di  and
      Schockaert, Steven",
    booktitle = "Proceedings of the 31st International Conference on Computational Linguistics",
    month = jan,
    year = "2025",
    address = "Abu Dhabi, UAE",
    publisher = "Association for Computational Linguistics",
    url = "https://aclanthology.org/2025.coling-main.266/",
    pages = "3936--3951",
}

@inproceedings{ilinykh-etal-2025-coreference,
    title = "Coreference as an indicator of context scope in multimodal narrative",
    author = "Ilinykh, Nikolai  and
      Lappin, Shalom  and
      Sayeed, Asad B.  and
      Lo{\'a}iciga, Sharid",
    booktitle = "Proceedings of the Fourth Workshop on Generation, Evaluation and Metrics (GEM{\texttwosuperior})",
    month = jul,
    year = "2025",
    address = "Vienna, Austria and virtual meeting",
    publisher = "Association for Computational Linguistics",
    url = "https://aclanthology.org/2025.gem-1.67/",
    pages = "789--807",
    ISBN = "979-8-89176-261-9",
}

@inproceedings{zamaraeva-etal-2025-comparing,
    title = "Comparing {LLM}-generated and human-authored news text using formal syntactic theory",
    author = "Zamaraeva, Olga  and
      Flickinger, Dan  and
      Bond, Francis  and
      G{\'o}mez-Rodr{\'i}guez, Carlos",
    editor = "Che, Wanxiang  and
      Nabende, Joyce  and
      Shutova, Ekaterina  and
      Pilehvar, Mohammad Taher",
    booktitle = "Proceedings of the 63rd Annual Meeting of the Association for Computational Linguistics (Volume 1: Long Papers)",
    month = jul,
    year = "2025",
    address = "Vienna, Austria",
    publisher = "Association for Computational Linguistics",
    url = "https://aclanthology.org/2025.acl-long.443/",
    doi = "10.18653/v1/2025.acl-long.443",
    pages = "9041--9060",
    ISBN = "979-8-89176-251-0"
}

@inproceedings{kasai-etal-2022-transparent,
    title = "Transparent Human Evaluation for Image Captioning",
    author = "Kasai, Jungo  and
      Sakaguchi, Keisuke  and
      Dunagan, Lavinia  and
      Morrison, Jacob  and
      Le Bras, Ronan  and
      Choi, Yejin  and
      Smith, Noah A.",
    editor = "Carpuat, Marine  and
      de Marneffe, Marie-Catherine  and
      Meza Ruiz, Ivan Vladimir",
    booktitle = "Proceedings of the 2022 Conference of the North American Chapter of the Association for Computational Linguistics: Human Language Technologies",
    month = jul,
    year = "2022",
    address = "Seattle, United States",
    publisher = "Association for Computational Linguistics",
    url = "https://aclanthology.org/2022.naacl-main.254/",
    doi = "10.18653/v1/2022.naacl-main.254",
    pages = "3464--3478"
}

@article{hong-etal-2023-visual-writing,
    title = "Visual Writing Prompts: Character-Grounded Story Generation with Curated Image Sequences",
    author = "Hong, Xudong  and
      Sayeed, Asad  and
      Mehra, Khushboo  and
      Demberg, Vera  and
      Schiele, Bernt",
    journal = "Transactions of the Association for Computational Linguistics",
    volume = "11",
    year = "2023",
    address = "Cambridge, MA",
    publisher = "MIT Press",
    url = "https://aclanthology.org/2023.tacl-1.33/",
    doi = "10.1162/tacl_a_00553",
    pages = "565--581"
}

@book{HallidayHasan1976,
author = {Halliday, M. A. K. and Hasan, Ruqaiya},
title = {Cohesion in English},
publisher = {Longman},
year = {1976},
address = {London}
}

@inproceedings{surikuchi-etal-2023-groovist,
    title = "{GROOV}i{ST}: A Metric for Grounding Objects in Visual Storytelling",
    author = "Surikuchi, Aditya K  and
      Pezzelle, Sandro  and
      Fern{\'a}ndez, Raquel",
    editor = "Bouamor, Houda  and
      Pino, Juan  and
      Bali, Kalika",
    booktitle = "Proceedings of the 2023 Conference on Empirical Methods in Natural Language Processing",
    month = dec,
    year = "2023",
    address = "Singapore",
    publisher = "Association for Computational Linguistics",
    url = "https://aclanthology.org/2023.emnlp-main.202/",
    doi = "10.18653/v1/2023.emnlp-main.202",
    pages = "3331--3339"
}

@misc{bai2025qwen3vltechnicalreport,
      title={Qwen3-VL Technical Report}, 
      author={Shuai Bai and Yuxuan Cai and Ruizhe Chen and Keqin Chen and Xionghui Chen and Zesen Cheng and Lianghao Deng and Wei Ding and Chang Gao and Chunjiang Ge and Wenbin Ge and Zhifang Guo and Qidong Huang and Jie Huang and Fei Huang and Binyuan Hui and Shutong Jiang and Zhaohai Li and Mingsheng Li and Mei Li and Kaixin Li and Zicheng Lin and Junyang Lin and Xuejing Liu and Jiawei Liu and Chenglong Liu and Yang Liu and Dayiheng Liu and Shixuan Liu and Dunjie Lu and Ruilin Luo and Chenxu Lv and Rui Men and Lingchen Meng and Xuancheng Ren and Xingzhang Ren and Sibo Song and Yuchong Sun and Jun Tang and Jianhong Tu and Jianqiang Wan and Peng Wang and Pengfei Wang and Qiuyue Wang and Yuxuan Wang and Tianbao Xie and Yiheng Xu and Haiyang Xu and Jin Xu and Zhibo Yang and Mingkun Yang and Jianxin Yang and An Yang and Bowen Yu and Fei Zhang and Hang Zhang and Xi Zhang and Bo Zheng and Humen Zhong and Jingren Zhou and Fan Zhou and Jing Zhou and Yuanzhi Zhu and Ke Zhu},
      year={2025},
      eprint={2511.21631},
      archivePrefix={arXiv},
      primaryClass={cs.CV},
      url={https://arxiv.org/abs/2511.21631}, 
}

@article{munoz-ortiz_contrasting_2024,
	title = {Contrasting {Linguistic} {Patterns} in {Human} and {LLM}-{Generated} {News} {Text}},
	volume = {57},
	issn = {1573-7462},
	url = {https://doi.org/10.1007/s10462-024-10903-2},
	doi = {10.1007/s10462-024-10903-2},
	number = {10},
	journal = {Artificial Intelligence Review},
	author = {Muñoz-Ortiz, Alberto and Gómez-Rodríguez, Carlos and Vilares, David},
	month = aug,
	year = {2024},
	pages = {265},
}

@article{jiang2025-artificial-hivemind,
  author       = {Liwei Jiang and
                  Yuanjun Chai and
                  Margaret Li and
                  Mickel Liu and
                  Raymond Fok and
                  Nouha Dziri and
                  Yulia Tsvetkov and
                  Maarten Sap and
                  Alon Albalak and
                  Yejin Choi},
  title        = {Artificial Hivemind: The Open-Ended Homogeneity of Language Models
                  (and Beyond)},
  journal      = {CoRR},
  volume       = {abs/2510.22954},
  year         = {2025},
  url          = {https://doi.org/10.48550/arXiv.2510.22954},
  doi          = {10.48550/ARXIV.2510.22954},
  eprinttype    = {arXiv},
  eprint       = {2510.22954},
  bibsource    = {dblp computer science bibliography, https://dblp.org}
}

@article{grootendorst2022bertopic,
  title={BERTopic: Neural topic modeling with a class-based TF-IDF procedure},
  author={Grootendorst, Maarten},
  journal={arXiv preprint arXiv:2203.05794},
  year={2022}
}

@article{10.1162/tacl_a_00543,
    author = {Bohnet, Bernd and Alberti, Chris and Collins, Michael},
    title = {Coreference Resolution through a seq2seq Transition-Based System},
    journal = {Transactions of the Association for Computational Linguistics},
    volume = {11},
    pages = {212-226},
    year = {2023},
    month = {03},
    issn = {2307-387X},
    doi = {10.1162/tacl_a_00543},
    url = {https://doi.org/10.1162/tacl_a_00543},
    eprint = {https://direct.mit.edu/tacl/article-pdf/doi/10.1162/tacl_a_00543/2074888/tacl_a_00543.pdf},
}

@inproceedings{porada-etal-2024-controlled-reevaluation,
    title = "A Controlled Reevaluation of Coreference Resolution Models",
    author = "Porada, Ian  and
      Zou, Xiyuan  and
      Cheung, Jackie Chi Kit",
    booktitle = "Proceedings of the 2024 Joint International Conference on Computational Linguistics, Language Resources and Evaluation (LREC-COLING 2024)",
    month = may,
    year = "2024",
    address = "Torino, Italia",
    publisher = "ELRA and ICCL",
    url = "https://aclanthology.org/2024.lrec-main.23",
    pages = "256--263",
}

@misc{yang2025qwen3technicalreport,
      title={Qwen3 Technical Report}, 
      author={An Yang and Anfeng Li and Baosong Yang and Beichen Zhang and Binyuan Hui and Bo Zheng and Bowen Yu and Chang Gao and Chengen Huang and Chenxu Lv and Chujie Zheng and Dayiheng Liu and Fan Zhou and Fei Huang and Feng Hu and Hao Ge and Haoran Wei and Huan Lin and Jialong Tang and Jian Yang and Jianhong Tu and Jianwei Zhang and Jianxin Yang and Jiaxi Yang and Jing Zhou and Jingren Zhou and Junyang Lin and Kai Dang and Keqin Bao and Kexin Yang and Le Yu and Lianghao Deng and Mei Li and Mingfeng Xue and Mingze Li and Pei Zhang and Peng Wang and Qin Zhu and Rui Men and Ruize Gao and Shixuan Liu and Shuang Luo and Tianhao Li and Tianyi Tang and Wenbiao Yin and Xingzhang Ren and Xinyu Wang and Xinyu Zhang and Xuancheng Ren and Yang Fan and Yang Su and Yichang Zhang and Yinger Zhang and Yu Wan and Yuqiong Liu and Zekun Wang and Zeyu Cui and Zhenru Zhang and Zhipeng Zhou and Zihan Qiu},
      year={2025},
      eprint={2505.09388},
      archivePrefix={arXiv},
      primaryClass={cs.CL},
      url={https://arxiv.org/abs/2505.09388}, 
}

@misc{zhu2025internvl3exploringadvancedtraining,
      title={InternVL3: Exploring Advanced Training and Test-Time Recipes for Open-Source Multimodal Models}, 
      author={Jinguo Zhu and Weiyun Wang and Zhe Chen and Zhaoyang Liu and Shenglong Ye and Lixin Gu and Hao Tian and Yuchen Duan and Weijie Su and Jie Shao and Zhangwei Gao and Erfei Cui and Xuehui Wang and Yue Cao and Yangzhou Liu and Xingguang Wei and Hongjie Zhang and Haomin Wang and Weiye Xu and Hao Li and Jiahao Wang and Nianchen Deng and Songze Li and Yinan He and Tan Jiang and Jiapeng Luo and Yi Wang and Conghui He and Botian Shi and Xingcheng Zhang and Wenqi Shao and Junjun He and Yingtong Xiong and Wenwen Qu and Peng Sun and Penglong Jiao and Han Lv and Lijun Wu and Kaipeng Zhang and Huipeng Deng and Jiaye Ge and Kai Chen and Limin Wang and Min Dou and Lewei Lu and Xizhou Zhu and Tong Lu and Dahua Lin and Yu Qiao and Jifeng Dai and Wenhai Wang},
      year={2025},
      eprint={2504.10479},
      archivePrefix={arXiv},
      primaryClass={cs.CV},
      url={https://arxiv.org/abs/2504.10479}, 
}

@misc{openai2024gpt4technicalreport,
      title={GPT-4 Technical Report}, 
      author={OpenAI and Josh Achiam and Steven Adler and Sandhini Agarwal and Lama Ahmad and Ilge Akkaya and Florencia Leoni Aleman and Diogo Almeida and Janko Altenschmidt and Sam Altman and Shyamal Anadkat and Red Avila and Igor Babuschkin and Suchir Balaji and Valerie Balcom and Paul Baltescu and Haiming Bao and Mohammad Bavarian and Jeff Belgum and Irwan Bello and Jake Berdine and Gabriel Bernadett-Shapiro and Christopher Berner and Lenny Bogdonoff and Oleg Boiko and Madelaine Boyd and Anna-Luisa Brakman and Greg Brockman and Tim Brooks and Miles Brundage and Kevin Button and Trevor Cai and Rosie Campbell and Andrew Cann and Brittany Carey and Chelsea Carlson and Rory Carmichael and Brooke Chan and Che Chang and Fotis Chantzis and Derek Chen and Sully Chen and Ruby Chen and Jason Chen and Mark Chen and Ben Chess and Chester Cho and Casey Chu and Hyung Won Chung and Dave Cummings and Jeremiah Currier and Yunxing Dai and Cory Decareaux and Thomas Degry and Noah Deutsch and Damien Deville and Arka Dhar and David Dohan and Steve Dowling and Sheila Dunning and Adrien Ecoffet and Atty Eleti and Tyna Eloundou and David Farhi and Liam Fedus and Niko Felix and Simón Posada Fishman and Juston Forte and Isabella Fulford and Leo Gao and Elie Georges and Christian Gibson and Vik Goel and Tarun Gogineni and Gabriel Goh and Rapha Gontijo-Lopes and Jonathan Gordon and Morgan Grafstein and Scott Gray and Ryan Greene and Joshua Gross and Shixiang Shane Gu and Yufei Guo and Chris Hallacy and Jesse Han and Jeff Harris and Yuchen He and Mike Heaton and Johannes Heidecke and Chris Hesse and Alan Hickey and Wade Hickey and Peter Hoeschele and Brandon Houghton and Kenny Hsu and Shengli Hu and Xin Hu and Joost Huizinga and Shantanu Jain and Shawn Jain and Joanne Jang and Angela Jiang and Roger Jiang and Haozhun Jin and Denny Jin and Shino Jomoto and Billie Jonn and Heewoo Jun and Tomer Kaftan and Łukasz Kaiser and Ali Kamali and Ingmar Kanitscheider and Nitish Shirish Keskar and Tabarak Khan and Logan Kilpatrick and Jong Wook Kim and Christina Kim and Yongjik Kim and Jan Hendrik Kirchner and Jamie Kiros and Matt Knight and Daniel Kokotajlo and Łukasz Kondraciuk and Andrew Kondrich and Aris Konstantinidis and Kyle Kosic and Gretchen Krueger and Vishal Kuo and Michael Lampe and Ikai Lan and Teddy Lee and Jan Leike and Jade Leung and Daniel Levy and Chak Ming Li and Rachel Lim and Molly Lin and Stephanie Lin and Mateusz Litwin and Theresa Lopez and Ryan Lowe and Patricia Lue and Anna Makanju and Kim Malfacini and Sam Manning and Todor Markov and Yaniv Markovski and Bianca Martin and Katie Mayer and Andrew Mayne and Bob McGrew and Scott Mayer McKinney and Christine McLeavey and Paul McMillan and Jake McNeil and David Medina and Aalok Mehta and Jacob Menick and Luke Metz and Andrey Mishchenko and Pamela Mishkin and Vinnie Monaco and Evan Morikawa and Daniel Mossing and Tong Mu and Mira Murati and Oleg Murk and David Mély and Ashvin Nair and Reiichiro Nakano and Rajeev Nayak and Arvind Neelakantan and Richard Ngo and Hyeonwoo Noh and Long Ouyang and Cullen O'Keefe and Jakub Pachocki and Alex Paino and Joe Palermo and Ashley Pantuliano and Giambattista Parascandolo and Joel Parish and Emy Parparita and Alex Passos and Mikhail Pavlov and Andrew Peng and Adam Perelman and Filipe de Avila Belbute Peres and Michael Petrov and Henrique Ponde de Oliveira Pinto and Michael and Pokorny and Michelle Pokrass and Vitchyr H. Pong and Tolly Powell and Alethea Power and Boris Power and Elizabeth Proehl and Raul Puri and Alec Radford and Jack Rae and Aditya Ramesh and Cameron Raymond and Francis Real and Kendra Rimbach and Carl Ross and Bob Rotsted and Henri Roussez and Nick Ryder and Mario Saltarelli and Ted Sanders and Shibani Santurkar and Girish Sastry and Heather Schmidt and David Schnurr and John Schulman and Daniel Selsam and Kyla Sheppard and Toki Sherbakov and Jessica Shieh and Sarah Shoker and Pranav Shyam and Szymon Sidor and Eric Sigler and Maddie Simens and Jordan Sitkin and Katarina Slama and Ian Sohl and Benjamin Sokolowsky and Yang Song and Natalie Staudacher and Felipe Petroski Such and Natalie Summers and Ilya Sutskever and Jie Tang and Nikolas Tezak and Madeleine B. Thompson and Phil Tillet and Amin Tootoonchian and Elizabeth Tseng and Preston Tuggle and Nick Turley and Jerry Tworek and Juan Felipe Cerón Uribe and Andrea Vallone and Arun Vijayvergiya and Chelsea Voss and Carroll Wainwright and Justin Jay Wang and Alvin Wang and Ben Wang and Jonathan Ward and Jason Wei and CJ Weinmann and Akila Welihinda and Peter Welinder and Jiayi Weng and Lilian Weng and Matt Wiethoff and Dave Willner and Clemens Winter and Samuel Wolrich and Hannah Wong and Lauren Workman and Sherwin Wu and Jeff Wu and Michael Wu and Kai Xiao and Tao Xu and Sarah Yoo and Kevin Yu and Qiming Yuan and Wojciech Zaremba and Rowan Zellers and Chong Zhang and Marvin Zhang and Shengjia Zhao and Tianhao Zheng and Juntang Zhuang and William Zhuk and Barret Zoph},
      year={2024},
      eprint={2303.08774},
      archivePrefix={arXiv},
      primaryClass={cs.CL},
      url={https://arxiv.org/abs/2303.08774}, 
}
\bibliographystyle{acl_natbib}

\appendix
\twocolumn

\section{Metric details}
\label{app:metric_details}

\paragraph{Notation}
A story has ordered segments ($s_1$, $\dots$, $s_N$), separated by \sepmark.
Note that each segment might have multiple sentences.
In implementation of character persistence and multimodal character grounding, we add a very small constant to each denominator to avoid division by zero in edge cases.

\subsection*{Topic switch}


Let $N$ be the number of segments in a story. Let $C$ be the number of adjacent segment pairs where the topic label changes. Then the topic switch score is:
\[
T=\frac{C}{N-1}.
\]

\noindent Here, $N-1$ is the number of adjacent segment pairs in the story. $T=0$ means no topic changes, while $T=1$ means the topic changes at every segment.

Given multiple topic granularities, we compute topic switch at several BERTopic settings by varying $\mathbf{nr\_topics}$ from 80 to 5 (step size 5), and then average the scores:
\[
T=\frac{1}{K}\sum_{j=1}^{K}T_j,
\]
where $K$ is the number of topic settings and $T_j$ is the topic switch score at setting $j$.

\subsection*{Character persistence}

For each matched character, we compute character continuity (ChC) as the fraction of adjacent segment pairs in which that character is mentioned in both segments.
Higher values mean stronger continuity across neighbouring segments.
We compute character spread (ChS) as the normalised distance between the first and the last segment where that character is mentioned.
Higher values mean that the character is re-mentioned more broadly across the story.

We combine continuity and spread into a single score:

\[
\mathrm{ChP}_c=\frac{\mathrm{ChC}_c}{\mathrm{ChS}_c}.
\]

\subsection*{Multimodal character grounding}

We define multimodal character grounding at the story level by combining two components: multimodal character continuity 
and GROOViST \citep{surikuchi-etal-2023-groovist}.
Similar to character persistence metric, we use character annotations from MovieNet \citep{huang2020movienetholisticdatasetmovie} linked with the output of the Link-Append system.
MCC captures whether character mentions in text are consistent with character presence in the image sequence.
GROOViST captures overall visual grounding quality of the story.
For story $i$, we compute

\[
\mathrm{MCG}_i=\frac{\mathrm{MCC}_i}{\mathrm{GV}_i}.
\]

\noindent Higher values indicate stronger multimodal character grounding, meaning stronger character-level multimodal continuity relative to visual grounding.

\subsection*{Narrative coherence score}

Let $M$ be the number of metrics in the framework (here, $M=5$), and let $x_1,\dots,x_M$ be the corresponding story-level metric values for one story (coreference, implicit discourse relation typology, topic switch, character persistence, and multimodal character grounding).

For $M$ component metrics $x_1,\dots,x_M$:
\[
\mathrm{NCS}^{\mathrm{arith}}=\frac{1}{M}\sum_{m=1}^{M}x_m,
\]
\[
\mathrm{NCS}^{\mathrm{geom}}=\sqrt[5]{x_1\,x_2\,x_3\,x_4\,x_5}.
\]

\noindent Here, $\mathrm{NCS}^{\mathrm{geom}}$ is more sensitive to imbalance across dimensions: low values on one or more components reduce the score more strongly than in the arithmetic mean.
Geometric mean is computed after adding a small constant for numerical stability.

\section{Model size and budget}
\label{app:model_budget}

We used A100 40 GB and A100 80 GB GPUs to run open-source models for visual story generation and coreference resolution.
In total, these tasks took approximately 24 hours.
Each model generated one story per sequence, using the same decoding settings (\(\texttt{temperature}=0.6\), \(\texttt{max\_tokens}=4096\)). 
Running the implicit discourse relation classifier took around 1 hour on a single NVIDIA GeForce RTX 3090 Ti GPU.
Closed-source models were prompted through their APIs.

\end{document}